\pdfoutput=1

\documentclass[11pt]{article}

\usepackage[]{ACL2023}

\usepackage{times}
\usepackage{latexsym}

\usepackage[T1]{fontenc}

\usepackage[utf8]{inputenc}

\usepackage{microtype}

\usepackage{inconsolata}

\usepackage{microtype}
\usepackage{lineno}
\usepackage{amsmath}
\usepackage{amssymb}
\usepackage{bbm}
\usepackage{multirow}
\usepackage{bm}
\usepackage{graphicx}
\usepackage{subfigure}
\graphicspath{{figures/}}

\usepackage{soul}
\usepackage{hyperref}

\title{TEPrompt: Task Enlightenment Prompt Learning \\ for Implicit Discourse Relation Recognition}

 \author{Wei Xiang \and Chao Liang \and Bang Wang \thanks{\quad Corresponding author: Bang Wang} \\
         School of Electronic Information and Communications, \\ 
         Huazhong University of Science and Technology, Wuhan, China \\
		\texttt{\{xiangwei, liangchao111, wangbang\}@hust.edu.cn}} 

\begin{document}
\maketitle
\begin{abstract}
Implicit Discourse Relation Recognition (IDRR) aims at classifying the relation sense between two arguments without an explicit connective. Recently, the ConnPrompt~\cite{Wei.X:et.al:2022:COLING} has leveraged the powerful prompt learning for IDRR based on the fusion of multi-prompt decisions from three different yet much similar connective prediction templates. Instead of multi-prompt ensembling, we propose to design auxiliary tasks with enlightened prompt learning for the IDRR task. Although an auxiliary task is not used to directly output final prediction, we argue that during the joint training some of its learned features can be useful to boost the main task. In light of such motivations, we propose a task enlightenment prompt learning model, called TEPrompt, to fuse learned features from three related tasks for IDRR. In particular, the TEPrompt contains three tasks, viz., Discourse Relation Recognition (DRR), Sense Semantics Classification (SSC) and Annotated Connective Prediction (ACP), each with a unique prompt template and an answer space. In the training phase, we jointly train three prompt learning tasks with shared argument representation. In the testing phase, we only take the DRR output with fused features as the final IDRR decision. Experiments with the same conditions have shown that the proposed TEPrompt outperforms the ConnPrompt. This can be attributed to the promoted decision features and language models benefited from joint-training of auxiliary tasks. 
\end{abstract}

\section{Introduction}
\begin{figure}[ht]
	\centering
	\includegraphics[width=\columnwidth, height=0.12\textwidth]{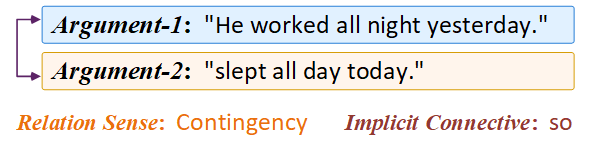}
	\caption{An example of implicit discourse relation annotation with manually inserted connective.}
	\label{Fig:Example}
\end{figure}

Implicit Discourse Relation Recognition (IDRR) is to detect and classify some latent relation in between a pair of text segments (called arguments) without an explicit connective~\cite{Wei.X:Bang.W:2022:CSUR}. Fig.~\ref{Fig:Example} illustrates an argument pair example with a \textsf{Contingency} relation in the Penn Discourse TreeBank (PDTB) corpus, and the implicit connective 'so' is inserted by annotators.
IDRR is of great importance for many downstream Natural Language Processing (NLP) applications, such as question answering~\cite{Liakata.M:et.al:2013:EMNLP}, machine translation~\cite{Guzman.F:et.al:2014:ACL}, summarization~\cite{Huang.Y.J:Kurohashi.S:2021:EACL}, and etc. However, due to the absence of an explicit connective, inferring discourse relations from the contextual semantics of arguments is still a challenging task.

\par
Conventional \textit{pre-train and fine-tuning} paradigm~\cite{Liu.P:et.al:2021:arXiv} designs sophisticated neural networks to encode the representation of argument pairs upon a Pre-trained Language Model (PLM) for relation classification~\cite{Chen.J:et.al:2016:ACL, Liu.Y:Li.S:2016:EMNLP, Ruan.H:et.al:2020:COLING,Li.X:et.al:2020:COLING,Liu.X:et.al:2020:IJCAI}.
On the one hand, these task-specific neural networks introduce some additional parameters that need to be trained by a large amount of labelled data. On the other hand, the task objective function is often not in accordance with that of the PLM, so that the PLM needs to be fine-tuned for solving downstream tasks, resulting in poor utilization of the encyclopedic linguistic knowledge embedded in the pre-training process.

\par
The recent ConnPrompt model~\cite{Wei.X:et.al:2022:COLING} has successfully applied the \textit{pre-train, prompt, and predict} paradigm, i.e. the so-called \textit{prompt learning}, in the IDRR task by transforming the IDRR as a connective-cloze task to predict an answer word and map it to a relation sense. The ConnPrompt has achieved the new state-of-the-art performance on the commonly used PDTB corpus~\cite{Webber.B:et.al:2019:UnivOfPenn}, however it designs three different yet much similar connective prediction templates which inserts the $\mathtt{[MASK]}$ token in between two arguments or at the beginning of one argument for answer prediction. Moreover, to fuse different prompt predictions, the ConnPrompt employs a simple majority voting decision fusing as for final relation sense prediction.

\par
Instead of simple multi-prompt ensemble, we argue that some auxiliary prompt tasks can be designed to enlighten the main prompt task with promoted decision features. For example, as the top relation labels in the PDTB corpus are those plain vocabulary words, we can design an auxiliary task to directly predict such label words from the PLM vocabulary. Furthermore, as the PDTB corpus also contains manually annotated implicit connectives, we can design another auxiliary task to directly predict an annotated connective. Although such auxiliary tasks are not necessarily used to output the final IDRR prediction, they can be jointly trained with the main task on a shared PLM, by which some features learned from the auxiliary tasks can be fused into the main task to promote its decision features for the final prediction. 

\par
Motivated from such considerations, we propose a \textit{Task Enlightenment Prompt Learning} (TEPrompt) model, where the main IDRR task can be enlightened from some auxiliary prompt tasks in terms of its promoted decision features via fusing auxiliary task features. Specifically, the TEPrompt contains a main prompt task: \textit{Discourse Relation Recognition} (DRR), and two auxiliary prompt tasks: \textit{Sense Semantics Classification} (SSC) and \textit{Annotated Connective Prediction} (ACP). We design each prompt task with a unique template and an answer space. We concatenate three prompt templates as an entire word sequence with two newly added special tokens $\mathtt{[Arg_1]}$ and $\mathtt{[Arg_2]}$ for shared argument representation, as the input of a PLM. In the training phase, we jointly train three prompt tasks upon one PLM model but with three different answer predictions as objective functions. In the testing phase, we only take the main prompt decision features yet promoted by fusing the features from the two auxiliary prompts to output the final IDRR decision.

\par
Experiment results have shown that our proposed TEPrompt outperforms the ConnPrompt with the same conditions and achieves the new state-of-the-art performance on the latest PDTB 3.0 corpus.

\begin{figure*}[ht]
	\centering
	\includegraphics[width=0.95\textwidth]{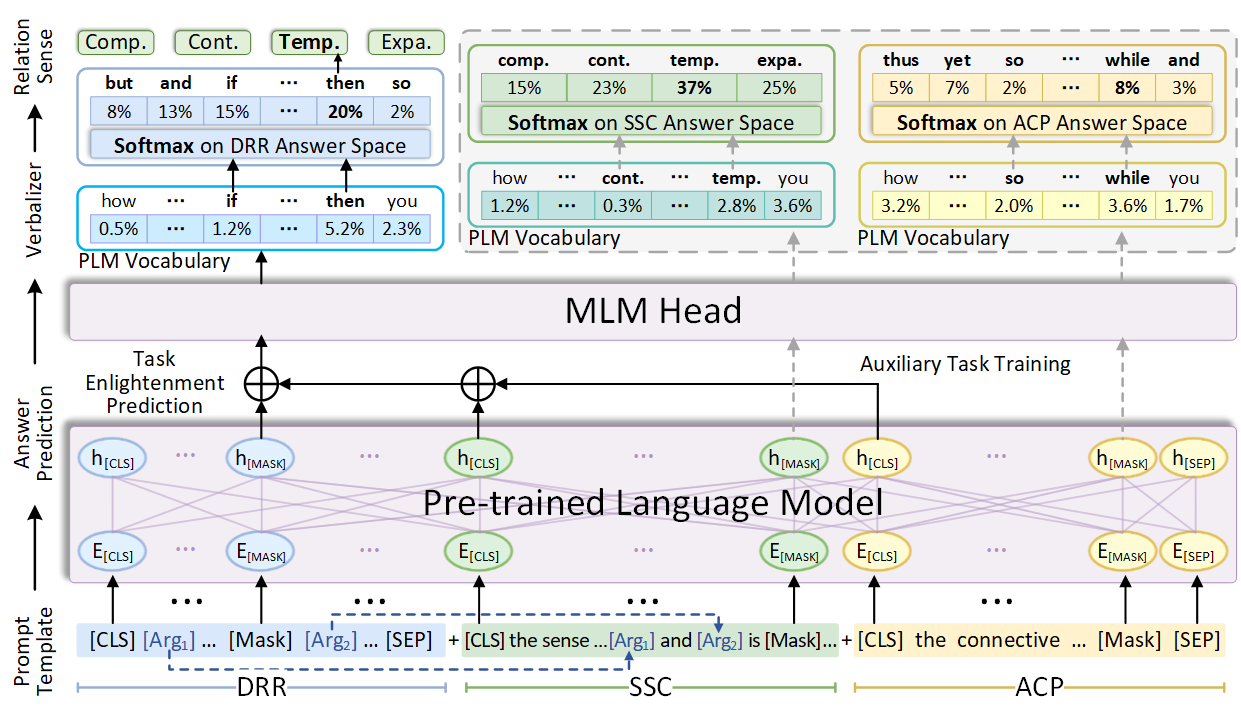}
	\caption{Illustration of our TEPrompt framework. It contains three modules of the prompt templatize, answer prediction and verbalizer for the main prompt task (DRR) and two auxiliary prompt tasks (SSC and ACP).}
	\label{Fig:Framework}
\end{figure*}

\section{Related Work}
\subsection{pre-train and fine-tuning paradigm}
Conventional pre-train and fine-tuning paradigm usually approaches the IDRR task as a classification problem, and the key is to design a sophisticated downstream neural network for argument representation learning~\cite{Zhang.B:et.al:2015:EMNLP,Rutherford.A:et.al:2017:EACL}. For example, the SCNN model~\cite{Zhang.B:et.al:2015:EMNLP} obtains each argument representation via a single convolution layer and concatenates two arguments' representations for relation classification. 
Some hybrid models have attempted to combine CNN, LSTM, graph convolutional networks and etc., for argument representation learning~\cite{Zhang.Y:et.al:2021:NAACL, Jiang.F:et.al:2021:EMNLP}. 

\par
Attention mechanisms have been widely used in neural model to unequally encode each word according to its importance for argument representation~\cite{Zhou.P:et.al:2016:ACL, Guo.F:et.al:2020:AAAI,Ruan.H:et.al:2020:COLING,Li.X:et.al:2020:COLING}.
For example, \citet{Zhou.P:et.al:2016:ACL} apply self-attention to weight a word according to its similarity to its belonging argument. 
\citet{Ruan.H:et.al:2020:COLING} propose a pipeline workflow to apply interactive attention after self-attention. \citet{Li.X:et.al:2020:COLING} use a penalty-based loss re-estimation method to regulate the attention learning.

\par
Word pair features have been exploited to capture interactions between arguments for representation learning~\cite{Chen.J:et.al:2016:AAAI, Chen.J:et.al:2016:ACL,  Wei.X:et.al:2022:ACL-Findings}. 
For example, \citet{Chen.J:et.al:2016:ACL} construct a relevance score word-pair interaction matrix based on a bilinear model~\cite{Jenatton.R:et.al:2012:NeurIPS} and a single layer neural model~\cite{Collobert.R:Weston.J:2008:ICML}. \citet{Wei.X:et.al:2022:ACL-Findings} propose an offset matrix network to encode word-pairs' offsets as linguistic evidence for argument representation.

\subsection{pre-train, prompt, and predict paradigm}
Recently, some large-scale PLMs have been proposed, such as the BERT~\cite{Devlin.J:et.al:2019:NAACL}, RoBERTa~\cite{Liu.Y:et.al:2019:arXiv}, T5~\cite{Raffel.C:et.al:2020:Jour.OfMachineLearn.Research}, and etc. The prompt learning has become a new paradigm for many NLP tasks, which uses the probability of text in PLMs to perform a prediction task, and has achieved promising results~\cite{Seoh.R:et.al:2021:EMNLP,Wang.C:et.al:2021:EMNLP,Ding.N:et.al:2021:arXiv}. For example, \citet{Seoh.R:et.al:2021:EMNLP} propose a cloze question prompt and a natural language inference prompt for aspect-based sentiment analysis.
\citet{Wang.C:et.al:2021:EMNLP} propose a transferable prompting framework to capture cross-task knowledge for few-shot text classification.
\citet{Ding.N:et.al:2021:arXiv} apply a cloze-style prompt learning on fine-grained entity typing in fully supervised, few-shot and zero-shot scenarios. 

\par
Some studies design appropriate prompts to reformulate an IDRR task for predicting discourse relations~\cite{Jiang.C:et.al:2021:PRICAI, Jiang.F:et.al:2021:EMNLP,Wei.X:et.al:2022:COLING}. ~\citet{Jiang.C:et.al:2021:PRICAI} use a masked PLM to generate a pseudo-connective for relation classification. \citet{Jiang.F:et.al:2021:EMNLP} utilize the PLM T5~\cite{Raffel.C:et.al:2020:Jour.OfMachineLearn.Research} to generate the target sentence which contains the meaning of discourse relations. \citet{Wei.X:et.al:2022:COLING} propose the ConnPrompt model with the new state-of-the-art performance, which reformulates the IDRR task as a connective-cloze task. They further use a majority voting decision fusion of the same task but with three much similar cloze templates for final relation sense prediction. 

\par
The proposed TEPrompt model fuses the learned features of two auxiliary prompt task to boost the main prompt tasks for relation prediction.

\section{The Proposed TEPrompt Model}
Fig.~\ref{Fig:Framework} presents our TEPrompt model, including three modules of prompt templatize, answer prediction and verbalizer for the main prompt task (DRR) and two auxiliary prompt tasks (SSC and ACP). The main DRR prompt task uses a kind of connective-cloze prompt to predict a manually selected answer words between two arguments, and map it to a relation sense; The SSC auxiliary prompt task describes and classifies the sense semantic between two arguments; While the ACP describes and predicts the implicit connective words.

\begin{figure*}[ht]
	\centering
	\includegraphics[width=0.8\textwidth]{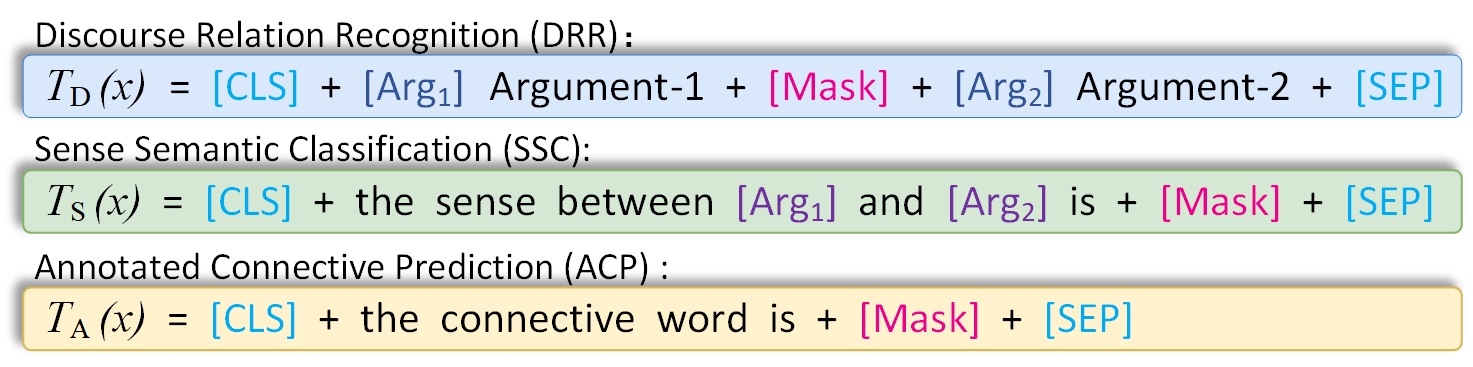}
	\caption{Illustration of our TEPrompt template, which is a concatenation of the three task templates.}
	\label{Fig:template}
\end{figure*}

\subsection{Prompt Templatize}
We first reformulate an input argument pair $x = (Arg_1; Arg_2)$ into a prompt template $T(x)$ by concatenating the main DRR prompt template with two auxiliary prompt templates: SSC and ACP, as the input of a PLM. Some PLM-specific tokens such as \texttt{[MASK]}, \texttt{[CLS]} and \texttt{[SEP]} are inserted in the prompt template; While the \texttt{[MASK]} tokens are added for the PLM to predict an answer word $v$, and the \texttt{[CLS]} and \texttt{[SEP]} tokens are used to indicate the beginning and ending of each prompt template, respectively. 

\par
Fig.~\ref{Fig:template} illustrates the three templates for our DRR, SSC and ACP task. We first use a kind of connective-cloze prompt template as the main DRR prompt template $T_{D}(x)$, in which argument-1 and argument-2 are concatenated as an entire word sequence, and the \texttt{[MASK]} token is inserted between two arguments. Besides, two newly added specific tokens $\mathtt{[Arg_1]}$ and $\mathtt{[Arg_2]}$ are inserted at the front of argument-1 and argument-2 to represent their semantics which are also shared in the SSC template.

\par
We also design two discrete prompt templates $T_S(x)$ and $T_A(x)$ for the auxiliary task SSC and ACP, respectively. The text of SSC template describes the sense semantics between argument-1 and argument-2; While the text of ACP template describes the implicit connective words. The \texttt{[MASK]} tokens are inserted at the end of SSC and ACP template for prediction. Note that in the SSC template, the specific tokens $\mathtt{[Arg_1]}$ and $\mathtt{[Arg_2]}$ are used to represent the semantics of argument-1 and argument-2, which are shared and trained with the main prompt task.

\subsection{Answer Prediction}
After the PLM, we obtain a hidden state $\mathbf{h}$ for each input token in the prompt templates, where $\mathbf{h} \in \mathbb{R}^{d_h}$ and $d_h$ is the dimension of the hidden state. We use $\mathbf{h}^{\textsl{DRR}}_{m}$, $\mathbf{h}^{\textsl{SSC}}_{m}$ and $\mathbf{h}^{\textsl{ACP}}_{m}$ to denote the hidden state of \texttt{[MASK]} tokens in the DRR, SSC and ACP template, respectively, which are used for the joint training of task enlightenment prompt learning; While the $\mathbf{h}^{\textsl{SSC}}_{c}$ and $\mathbf{h}^{\textsl{ACP}}_{c}$ are used to denote the hidden state of the \texttt{[CLS]} token in the SSC and ACP template, respectively, which are used for the feature fusion of auxiliary prompt tasks.

\par
To fuse the features of auxiliary prompt SSC and ACP into the main DRR task, we use the fusion gate mechanism to integrate their \texttt{[CLS]} representations into the \texttt{[MASK]} representation of the main DRR task, which is next used for the final answer word prediction. Specifically, we first use a fusion gate mechanism to integrate the \texttt{[CLS]} representations of SSC and ACP, the transition functions are computed as follows:
\begin{align} \label{}
	\mathbf{g}_c & = sigmoid(\mathbf{W}_c \mathbf{h}^{\textsl{SSP}}_{c} + \mathbf{U}_c \mathbf{h}^{\textsl{CEP}}_{c}), \\
	\mathbf{\tilde{h}}_c & = \mathbf{g}_c \odot \mathbf{h}^{\textsl{SSP}}_{c} + (1 - \mathbf{g}_c) \odot \mathbf{h}^{\textsl{CEP}}_{c},
\end{align}
where $\mathbf{W}_c \in \mathbb{R}^{d_h \times d_h}$, $\mathbf{U}_c \in \mathbb{R}^{d_h \times d_h}$ are learnable parameters and $\odot$ donates the element-wise product of vectors.

\par
With the fusion gate, we adaptively assign different importance to the features of SSC and ACP prompt task, and outputs $\mathbf{\tilde{h}}_c \in \mathbb{R}^{d_h}$ as the auxiliary prompt vector. We next use another fusion gate to integrate the auxiliary prompt vector $\mathbf{\tilde{h}}_c$ into the \texttt{[MASK]} hidden state of the main DRR prompt $\mathbf{h}^{\textsl{DRP}}_{m}$ for the final answer prediction. The transition functions are:
\begin{align} \label{}
	\mathbf{g}_m & = sigmoid(\mathbf{W}_m \mathbf{h}^{\textsl{DRP}}_{m} + \mathbf{U}_m \mathbf{\tilde{h}}_{c}), \\
	\mathbf{\tilde{h}}_m & = \mathbf{g}_m \odot \mathbf{h}^{\textsl{DRP}}_{m} + (1 - \mathbf{g}_m) \odot \mathbf{\tilde{h}}_{c},
\end{align}
where $\mathbf{W}_m \in \mathbb{R}^{d_h \times d_h}$, $\mathbf{U}_m \in \mathbb{R}^{d_h \times d_h}$ are learnable parameters.

\par
Finally, the Masked Language Model (MLM) classifier of the PLM uses the fused hidden state $\mathbf{\tilde{h}}_m$ to estimates the probability of each word in its vocabulary $V$ for the \texttt{[MASK]} token of the DRR task as follows:
\begin{align}
	P_D(\texttt{[MASK]}_\textsl{DRP} = v_d \in V \  | \  T(x)).
\end{align}
Note that, the MLM classifier also estimates an answer word probability $P_S$ and $P_A$ for the \texttt{[MASK]} token of the auxiliary prompt task SSC and ACP without feature fusion in the joint training.

\subsection{Verbalizer}
We define a discrete answer space for the DRR, SSC and ACP prompt task, respectively, which are all subsets of the PLM vocabulary. Specifically, we use sixteen manually selected answer words as the answer space $V_d$ of the DRR, the same as that of ConnPrompt~\cite{Wei.X:et.al:2022:COLING}. Besides, we use four top-level sense labels in the PDTB corpus as the SSC answer space, $V_s = \{\textsf{Comparison}$, $\textsf{Contingency}$, $\textsf{Expansion}$, $\textsf{Temporal}\}$, and we use the 174 manually annotated implicit connectives in the PDTB corpus as the ACP answer space $V_c$ of ACP. We note that the answer space of DRR is next mapped to a relation sense in verbalizer process, while the answer space of SSC and ACP are only used in the auxiliary task training.

\begin{table}[ht]
	\resizebox{\columnwidth}{!}{
		\renewcommand\arraystretch{1.1}
		\begin{tabular}{l|l}
			\hline
			{Relation Sense}                                  & \multicolumn{1}{l}{{Answer words}} \\
			\hline
			Comparison                                               & \textit{similarly}, \textit{but}, \textit{however}, \textit{although}         \\			
			Contingency                                              & \textit{for}, \textit{if}, \textit{because}, \textit{so}                      \\
			Expansion                                                & \textit{instead}, \textit{by}, \textit{thereby}, \textit{specifically}, \textit{and}   \\
			Temporal                                                 & \textit{simultaneously}, \textit{previously}, \textit{then}          \\
			\hline
	\end{tabular}}
	\caption{Answer space of the DRR prompt and the connection to the top-level class discourse relation sense labels in the PDTB corpus.}
	\label{Tab:Answer}
\end{table}

\par
After answer prediction, a softmax layer is applied on the prediction scores of our pre-defined answer space to normalize them into probabilities:
\begin{align}
	P(v \in V \ |\ T(x)) = \dfrac{e^{p_{v_i}}}{\sum_{j=1}^{n}e^{p_{v_j}}}.
\end{align}
Then, the predicted answer word of DRR is projected into a unique discourse relation sense based on the pre-defined connection regulation. Table~\ref{Tab:Answer} presents the verbalizer connection from the answer word to the PDTB discourse relation sense labels.

\subsection{Training and Prediction}
In the training phase, we tune the PLM parameters based on the DRR, SSC and ACP prompt task jointly to fuse their learned features. We compute a cross entropy loss for the DRR loss $L_d$, SSC loss $L_s$ and ACP loss $L_c$, respectively.
\begin{align} \label{}
	J(\theta) = -\frac{1}{K} \sum\limits_{k=1}^{K} \mathbf{y}^{(k)} \log(\mathbf{\hat{y}}^{(k)}) + \lambda \Vert \theta \Vert^2,
\end{align}
where $\mathbf{y}^{(k)}$ and $\mathbf{\hat{y}}^{(k)}$ are the answer label and predicted answer of the $k$-th training instance respectively. $\lambda$ and $\theta$ are the regularization hyper-parameters. We use the AdamW optimizer~\cite{Loshchilov.I:Hutter.F:2019:ICLR} with $L2$ regularization for model training.
The cost function of our TEPrompt is optimized as follows:
\begin{align} \label{}
	L = L_d + \beta L_s + \gamma L_c,
\end{align}
where $\beta$ and $\gamma$ are weight coefficients to balance the importance of the SSC loss and ACP loss.

\section{Experiment Setting}
In this section, we present our experiment settings, including the dataset, PLMs, competitors, and parameter settings.

\par
\textbf{The PDTB 3.0 Dataset}:
Our experiments are conducted on the Penn Discourse TreeBank (PDTB) 3.0 corpus~\footnote{We have purchased the PDTB 3.0 liscence for experiments.}~\cite{Webber.B:et.al:2019:UnivOfPenn}, which contains more than one million words of English texts from the Wall Street Journal. Following the conventional data splitting, we use sections 2-20 as the full training set, sections 21-22 as the testing set and 0-1 as the development set~\cite{Ji.Y:Eisenstein.J:2015:Trans.OfACL}. Our experiments are conducted on the four top-level classes of relation sense, including \textsf{Comparison}, \textsf{Contingency}, \textsf{Expansion}, \textsf{Temporal}. Table~\ref{Tab:Cropus} presents the dataset statistics.
\begin{table}[ht] 
	\centering
	\resizebox{0.68\columnwidth}{!}{
		\renewcommand\arraystretch{1.1}
		\begin{tabular}{l|lll}
			\hline
			Relation & Train & Dev. & Test \\ \hline
			Expansion    & 8645  & 748  & 643  \\
			Comparison    & 1937  & 190  & 154  \\
			Contingency    & 5916  & 579  & 529  \\
			Temporal    & 1447  & 136  & 148  \\ \hline
			Total    & 17945 & 1653 & 1474 \\ \hline
	\end{tabular}}
	\caption{Statistics of implicit discourse relation instances in PDTB 3.0 with four top-level relation senses.}
	\label{Tab:Cropus}
\end{table}

\par
\textbf{Pre-trained Language Models}: 
We use two of the most representative masked pre-trained language models (PLM) for comparison: \textbf{BERT}~\cite{Devlin.J:et.al:2019:NAACL} is the first Transformer-based large-scale pre-trained PLM proposed by Google~\footnote{https://github.com/google-research/bert}, which is pre-trained using a \textit{cloze task} and a \textit{next sentence prediction} task; \textbf{RoBERTa}~\cite{Liu.Y:et.al:2019:arXiv} is a BERT-enhanced PLM proposed by Facebook~\footnote{https://github.com/pytorch/fairseq/}, which removes the next sentence prediction objective and is pre-trained on a much larger dataset with some modified key hyper-parameters.
	
\par
\textbf{Competitors}:
We compare our TEPrompt with the following advanced models:
\par 
$\bullet$ \textsf{DAGRN}~\cite{Chen.J:et.al:2016:ACL} encodes word-pair interactions by a neural tensor network.
\par
$\bullet$ \textsf{NNMA}~\cite{Liu.Y:Li.S:2016:EMNLP} combines two arguments' representations for stacked interactive attentions.
\par
$\bullet$ \textsf{IPAL}~\cite{Ruan.H:et.al:2020:COLING} propagates self-attention into interactive attention by a cross-coupled network.
\par
$\bullet$ \textsf{PLR}~\cite{Li.X:et.al:2020:COLING} uses a penalty-based loss re-estimation to regulate the attention learning.
\par
$\bullet$ \textsf{BMGF}~\cite{Liu.X:et.al:2020:IJCAI} combines bilateral multi-perspective matching and global information fusion to learn a contextualized representation.
\par
$\bullet$ \textsf{MANF}~\cite{Wei.X:et.al:2022:ACL-Findings} encodes two kinds of attentive representation for arguments and fuses them with the word-pairs features.
\par
$\bullet$ \textsf{ConnPrompt}~\cite{Wei.X:et.al:2022:COLING} applies the prompt learning for IDRR based on the fusion of multi-prompt decisions.

\par
\textbf{Parameter Setting}:
We implement the PLM models with 768-dimension provided by HuggingFace transformers~\footnote{https://github.com/huggingface/transformers}~\cite{Wolf.T:et.al:2020:EMNLP}, and run PyTorch~\footnote{pytorch.org} framework with CUDA on NVIDIA GTX 3090 Ti GPUs. The maximum length of our TEPrompt template is set to 150 tokens, in which the maximum length of arguments are 70 tokens. 
We set the mini-batch size to 32, the learning rate to 1e-5, the weight coefficients $\beta$ and $\gamma$ to 0.3 and 0.4 respectively, and all trainable parameters are randomly initialized from normal distributions. We release the code at: https://github.com/HustMinsLab/TEPrompt.

\section{Result and Analysis}

\subsection{Overall Result}
Table~\ref{Tab:Overall Result} compares the overall performance between our TEPrompt and the competitors. We implement a four-way classification on the top-level relation sense of the PDTB dataset and adopt the commonly used macro $F1$ score and accuracy (Acc) as performance metrics. 

\par
We note that the competitors in the first group all use the pre-train and fine-tuning paradigm; While our TEPrompt and the ConnPrompt use the pre-train, prompt, and predict paradigm, i.e. the prompt learning. Besides, the first two competitors both use a kind of distributed and static word embeddings: Word2vec and Glove; while the others use Transformer-based PLM models: BERT and RoBERTa.

\par
The first observation is that the \textsf{DAGRN} and \textsf{NNMA} cannot outperform the other competitors. This is not unexpected, as the others employ the more advanced dynamic PLMs pre-trained with deeper neural networks and larger scale of parameters, which have been proven more effective for many downstream NLP tasks~\cite{Devlin.J:et.al:2019:NAACL,Liu.Y:et.al:2019:arXiv}. The gaps between large PLM fine-tuning and static embedding for representation learning also have a certain impact on the performance of the IDRR task.

\par
The second observation is that our TEPrompt and the ConnPrompt adopting the prompt learning paradigm can significantly outperform the other competitors in terms of much higher macro F1 score (8\%+) and Acc(5\%+). The outstanding performance can be attributed to the task transformation of connective-cloze prediction into the training of PLMs, other than designing a task-specific model upon PLM, by which the model can better enjoy the encyclopedic linguistic knowledge embedded in a PLM during the model training.

\par
Finally, our TEPrompt achieves better performance than the ConnPrompt with the same PLM and outperforms all the other models in both higher macro F1 score and accuracy. Similar results can also be observed in the binary classification (i.e. one-versus-others) of implicit discourse relation recognition, in Table~\ref{Tab:Binary Classification}. We attribute the outstanding performance of our TEPrompt to the use of auxiliary tasks for enlightenment prompt learning, by which the jointly trained features of auxiliary SSC and ACP prompt task can be well fused into the main DRR task to improve the final answer prediction. This will be further analyzed in our ablation study.

\begin{table}[t]
	\centering
	\resizebox{\columnwidth}{!}{
		\renewcommand\arraystretch{1.2}
		\begin{tabular}{l|c|cc}
			\hline
			Model                      &PLM         & Acc (\%)           & F1 (\%)            \\ \hline
			DAGRN (ACL, 2016)          &Word2vec    & 57.33              & 45.11              \\
			NNMA (EMNLP, 2016)         &Glove       & 57.67              & 46.13              \\
			IPAL (COLING, 2020)        &BERT        & 57.33              & 51.69              \\
			PLR (COLING, 2020)         &BERT        & 63.84              & 55.74              \\
			BMGF (IJCAI, 2020)         &RoBERTa     & 69.95              & 62.31              \\ 
			MANF (ACL-Findings, 2022)  &BERT        & 64.04              & 56.63              \\ \hline
			ConnPrompt (COLING, 2022)  &BERT        & 69.67              & 64.00              \\ 
			\textbf{Our TEPrompt}      &BERT        & 70.08              & 65.12              \\ \hline
			ConnPrompt (COLING, 2022)  &RoBERTa     & \underline{75.17}  & \underline{70.88}  \\ 
			\textbf{Our TEPrompt}      &RoBERTa     & \textbf{75.51}     &\textbf{ 72.26}     \\ \hline     
	\end{tabular}}
	\caption{Comparison of overall results on the PDTB.}
	\label{Tab:Overall Result}
\end{table}

\begin{table}[ht] 
	\centering
	\resizebox{0.99\columnwidth}{!}{
		\renewcommand\arraystretch{1.1}
		\begin{tabular}{l|llll}
			\hline
			Model               & Expa.             & Comp.             & Cont.             & Temp. \\ \hline
			DAGRN (ACL, 2016)   & 64.71             & 27.34             & 62.56             & 38.91    \\
			NNMA (EMNLP, 2016)  & 65.10             & 29.15             & 63.33             & 41.03    \\
			DERM (COLING, 2018) & 64.96             & 41.71             & 67.73             & 46.73    \\
			IPAL (COLING, 2020) & 66.86             & 37.31             & 66.40             & 41.25    \\
			PLR (COLING, 2020)  & 69.33             & 35.16             & 66.97             & 43.40    \\
			BMGF (IJCAI, 2020)  & \underline{72.61} & \underline{50.85} & \underline{72.42} & \underline{45.23}    \\
			MANF (ACL, 2022)    & {70.00}           & {35.83}           & {66.77}           & 40.22    \\
			\hline
			\textbf{Our TEPrompt} & \textbf{77.34}    & \textbf{53.42}    & \textbf{77.98}    & \textbf{53.55}  \\
			\hline
	\end{tabular}}
	\caption{Comparison  of binary classification results on the PDTB (F1 score \%). We have reproduced some of the competitors on PDTB 3.0 for fair comparison.}
	\label{Tab:Binary Classification}
\end{table}

\subsection{Ablation Study}
To examine the effectiveness of different prompt tasks, we design the following ablation studies. 

\par
$\bullet$ \textsf{Prompt-SSC} is only the SSC prompt concatenating argument-1 and argument-2 in front, without the DRR and ACP task.
\par
$\bullet$ \textsf{TEPrompt-SSC} combines the SCC prompt with DRR and ACP, and only uses the predicted answer of SSC for relation sense mapping.
\par
$\bullet$ \textsf{Prompt-ACP} is only the ACP prompt concatenating argument-1 and argument-2 in front, without the DRR and SSC.
\par
$\bullet$ \textsf{TEPrompt-ACP} combines the ACP prompt with the DRR and SSC, and uses the predicted answer of ACP for relation sense mapping~\footnote{Note that some implicit connectives correspond to multiple relation senses, we choose the one with the highest frequency in the training data as the prediction relation sense.}.
\par
$\bullet$ \textsf{Prompt-DRR} is only the DRR prompt without the auxiliary prompt SSC and ACP.
\par
$\bullet$ \textsf{TEPrompt w/o Gate} is our task enlightenment prompt model without fusion mechanisms.

\par
Table~\ref{Tab:Ablation} compares the results of our ablation study models with both single-prompt and multi-prompt ConnPrompt.

\begin{table}[htbp]
	\centering
	\resizebox{\columnwidth}{!}{
		\renewcommand\arraystretch{1.2}
		\begin{tabular}{l|cc|cc}
			\hline
			\multirow{2}{*}{PLM} & \multicolumn{2}{c|}{BERT} & \multicolumn{2}{c}{RoBERTa}   \\ \cline{2-5}
			& Acc (\%)    & F1 (\%)        & Acc (\%)    & F1 (\%)              \\ \hline
			ConnPrompt-1            & 69.74       & 63.95          & 74.36       & 69.91                 \\
			ConnPrompt-2            & 69.34       & 63.69          & 73.61       & 69.63                 \\
			ConnPrompt-3            & 67.64       & 62.65          & 73.54       & 69.00                \\ 
			ConnPrompt-Multi        & 69.67       & 64.00          & 75.17       & 70.88                \\ \hline
			Prompt-SSC              & 67.37       & 60.64          & 70.62       & 66.09                 \\
			TEPrompt-SSC            & 67.64       & 62.73          & 74.22       & 69.93                \\ 
			Prompt-ACP              & 66.08       & 59.08          & 72.73       & 67.89                 \\
			TEPrompt-ACP            & 67.23       & 61.44          & 73.13       & 68.83                \\ 
            Prompt-DRR              & 69.54       & 63.00          & 74.02       & 69.77                 \\
			\hline
			TEPrompt w/o Gate   & 68.32       & 63.48          & 75.03       & 70.58                \\
           \hline
           TEPrompt            & \textbf{70.08}       & \textbf{65.12}          & \textbf{75.51}       & \textbf{72.26}                \\
			\hline
	\end{tabular}}
	\caption{Results of ablation study on the PDTB corpus.}
	\label{Tab:Ablation}
\end{table}

\par
\textbf{Task enlightenment prompt:} We can observe that the \textsf{Prompt-DRR} has comparable performance to each single-ConnPrompt, viz. ConnPrompt-1/2/3. This is not unexpected. All the three single-ConnPrompts are with the same connective-cloze prompt model, and the only difference is the location of the cloze-mask in each template; While the \textsf{Prompt-DRR} is with the same connective-cloze prompt model and answer space as a single-ConnPrompt. The \textsf{ConnPrompt-Multi} uses multi-prompt majority voting and outperforms any of the single-ConnPrompt; While the \textsf{TEPrompt} designs two auxiliary tasks to augment the main task and outperforms both \textsf{Prompt-DRR} and \textsf{ConnPrompt-Multi}, which validates the effectiveness of our task enlightenment prompt learning via fusing features from both main and auxiliary prompt tasks by joint training.

\par
\textbf{Prompt ablation study:} Among the second group of prompt ablation models, it can be observed that the \textsf{Prompt-SSC} and \textsf{Prompt-ACP} cannot outperform the \textsf{Prompt-DRR}; While the \textsf{TEPrompt-SSC} and \textsf{TEPrompt-ACP} also cannot outperform the \textsf{TEPrompt}. Although both the SSC and ACP prompt model can each output the final prediction by mapping its predicted answer to a relation sense, however, their objectives are not completely in accordance with the IDRR task.
The SCC prompt is designed to classify sense semantics; While the ACP prompt aims at predicting manually annotated connectives. 
Furthermore, we can also observe that the \textsf{TEPrompt-SSC} and \textsf{TEPrompt-ACP} have achieved better performance than the \textsf{Prompt-SSC} and \textsf{Prompt-ACP}, respectively. This again validates our argument that fusing features from jointly trained auxiliary prompt tasks can be useful to boost the main prompt task prediction.

\begin{figure}[h]
	\centering
	\hspace{0.5mm}
	\subfigure[\label{SubFig:BERT} BERT ]
	{\includegraphics[width=0.9\columnwidth]{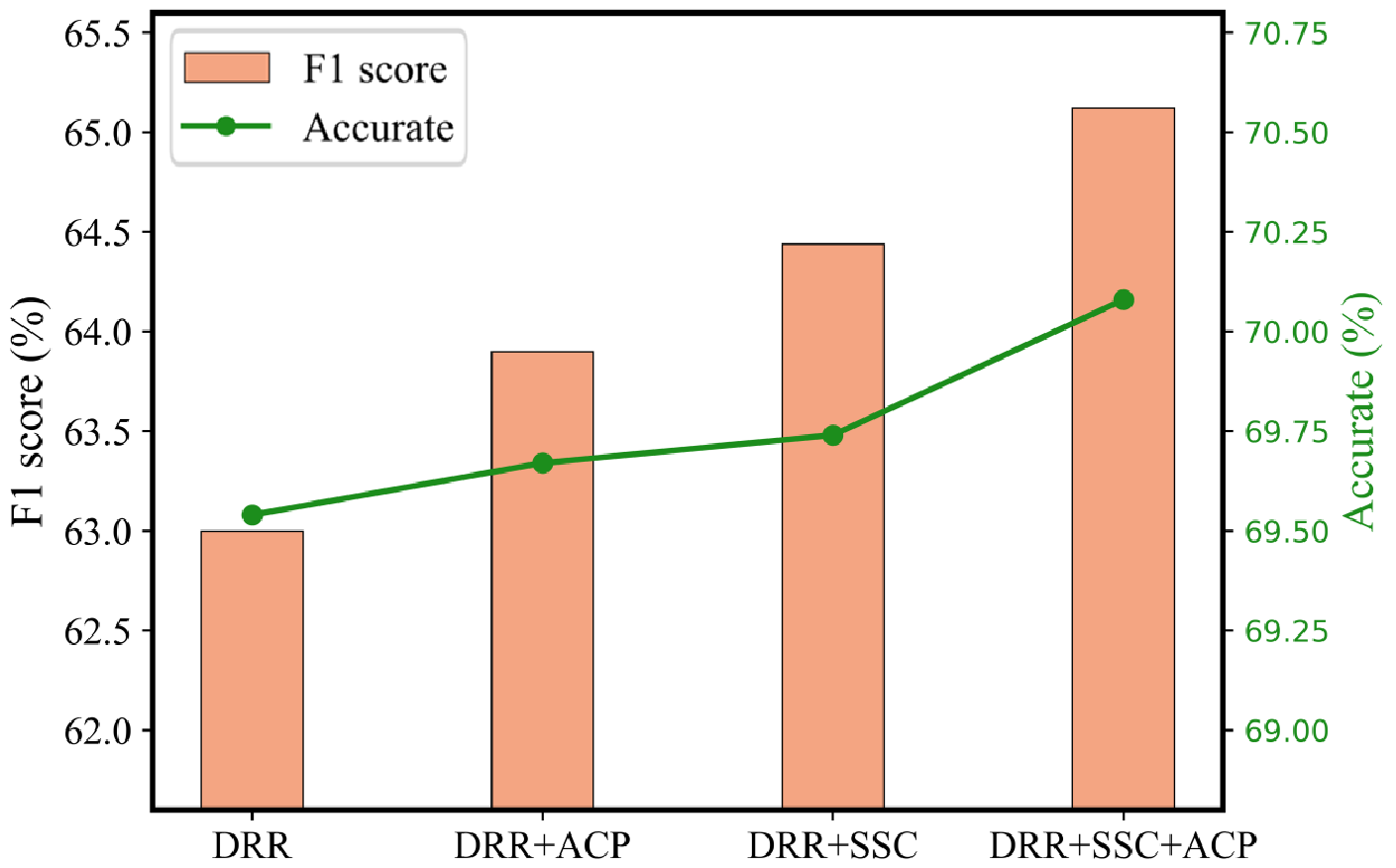}}
	\subfigure[\label{SubFig:RoBERTa} RoBERTa]
	{\includegraphics[width=0.88\columnwidth]{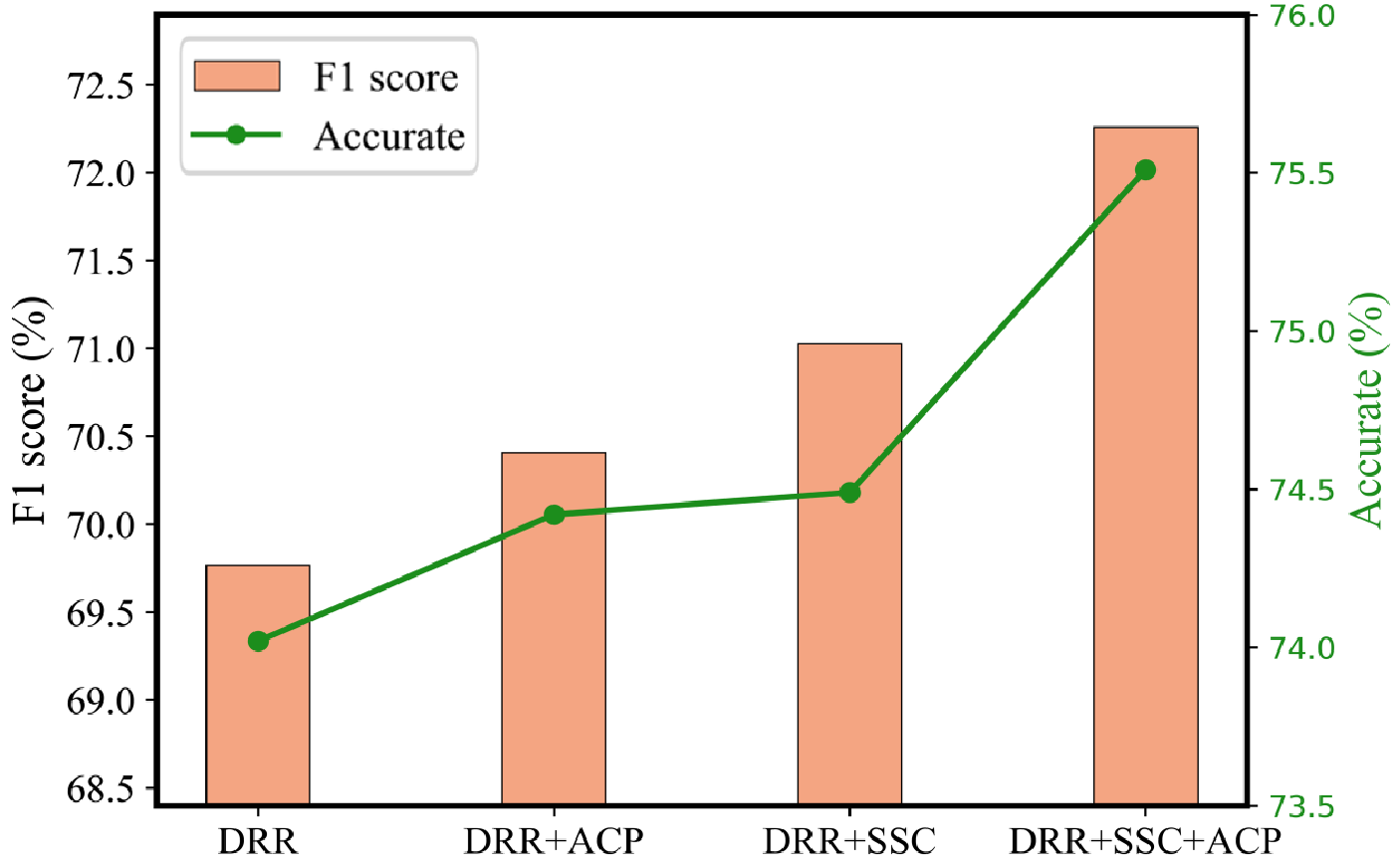}}                     
	\caption{Comparison of auxiliary prompt effections.}
	\label{Fig:Auxiliary}
\end{figure}

\begin{figure*}[ht]
	\centering
	\includegraphics[width=0.83\textwidth]{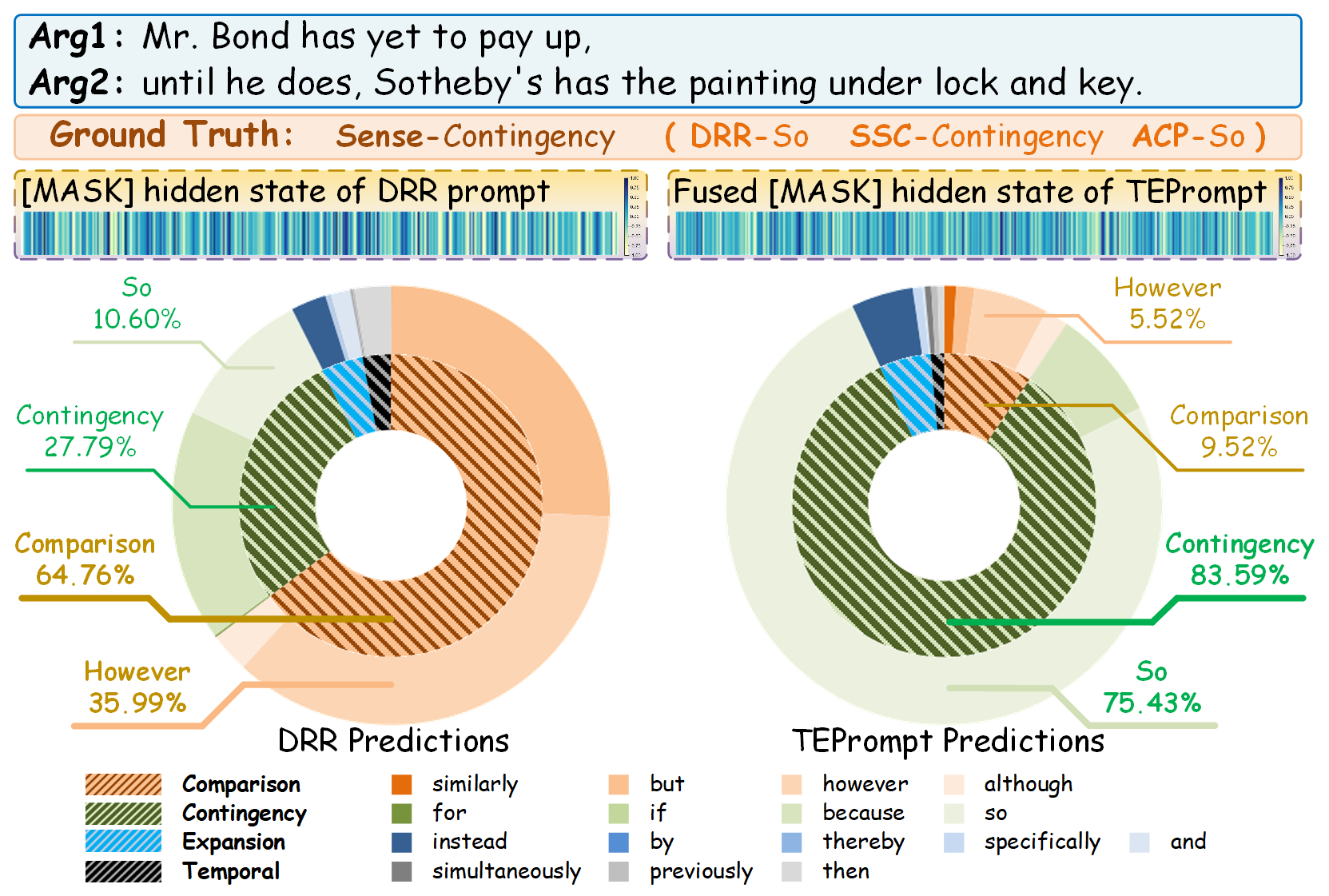}
	\caption{Visualization of the predicted answer words and relation sense for the DRR Prompt and TEPrompt.}
	\label{Fig:CaseStudy}
\end{figure*}

\par
\textbf{Gate Fusion Mechanism:} We also observe that the \textsf{TEPrompt w/o Gate} without gate fusion mechanism cannot outperform the full \textsf{TEPrompt} model, even it jointly trains a PLM as well as the MLM head with two auxiliary tasks. This indicates that the features learned from auxiliary tasks can indeed augment the main task prediction.

\par
\textbf{Auxiliary prompt effections:} To further investigate the task enlightenment effections, we design several combinations of individual prompt models: the \textsf{DRR} with the only main task, the \textsf{DRR+SSC} and \textsf{DRR+ACP} are the main task enlightened by only one auxiliary task, and \textsf{DRR+SSC+ACP} (viz., \textsf{TEPrompt}) is the main task enlightened by two auxiliary tasks.

\par
Fig.~\ref{Fig:Auxiliary} compares the performance of different auxiliary prompt ablation models. We can observe that both the SSC and ACP auxiliary task can help improving the performance of the main DRR task. This suggests that fusing either the sense semantics feature in training SSC or the annotated connective feature in training ACP (viz., the two \textsf{[CLS]} tokens) can help promoting the decision feature of the main DRR task (viz., the \textsf{[MASK]} token) to improve the IDRR prediction. Finally, our \textsf{TEPrompt} joint training with both SSC and ACP auxiliary prompts yields substantial improvements over all ablation models, again approving our arguments and design objectives.

\subsection{Case Study}
We use a case study to compare the TEPrompt and  the DRR prompt. Note that the DRR prompt can be regarded as the ConnPrompt using only one template yet without multi-prompt ensemble. Fig.~\ref{Fig:CaseStudy} visualizes the representation of the \texttt{[MASK]} token, as well as its prediction probability and classified relation sense by a pie chart. The \texttt{[MASK]} token representation of the TEPrompt is quite different from that of the DRR prompt, as the former also fuses two auxiliary prompt task features. Such feature fusion from auxiliary tasks may enlighten the main task to make correct predictions.

\par
It can be observed that the DRR prompt itself tends to predict a \textsf{Comparison} relation (64.76\%) corresponding to the answer word \textit{'however'} with the highest probability 35.99\%. After feature fusion, the TEPrompt can correctly recognize the \textsf{Contingency} relation (83.59\%) between the two arguments by predicting the answer word \textit{'so'} with a much higher probability 75.43\% than that of the DRR prompt prediction (10.60\%). We argue that such benefits from the adjustments of prediction probabilities can be attributed to the feature fusion of the two auxiliary prompt tasks.

\section{Concluding Remarks}
In this paper, we have argued a main prompt task can be enlightened by some auxiliary prompt tasks for performance improvements. For the IDRR task, we have proposed a TEPrompt, a task enlightenment prompt model that fuses learned features from our designed auxiliary SSC and ACP task into the decision features of the main DRR task. Since the three prompt tasks are trained jointly, the learned auxiliary task features in the training phase can help promoting the main task decision feature and improving the final relation prediction in the testing phase. Experiment results and ablation studies have validated the effectiveness of our arguments and design objectives in terms of improved state-of-the-art IDRR performance. 

\par
In our future work, we shall investigate other types of auxiliary tasks for the IDRR task as well as the applicability of such task enlightenment prompt learning  for other NLP tasks. 

\section*{Limitations}
The two auxiliary prompt tasks are closely related to the PDTB corpus, as the top-level relation sense labels are those plain vocabulary words and the PDTB provides manually annotated connectives. 

%


\section*{Acknowledgements}
This work is supported in part by National Natural Science Foundation of China (Grant No: 62172167). The computation is completed in the HPC Platform of Huazhong University of Science and Technology.

\bibliography{mybibfile}
\bibliographystyle{acl_natbib}

%

\end{document}